\documentclass[journal]{IEEEtran}
\usepackage{amssymb}
\usepackage{array}
\usepackage{amsmath}
\usepackage{multirow}
\ifCLASSINFOpdf
   \usepackage[pdftex]{graphicx}
   \DeclareGraphicsExtensions{.pdf,.jpeg,.png}
\else
\fi

\usepackage{xcolor}
\begin{document}
%
\title{Heterogeneous Strategy Particle Swarm Optimization}
%
%
%

\author{Wen-Bo~Du,
        Wen~Ying,
        Gang~Yan,
        Yan-Bo~Zhu,
        and~Xian-Bin~Cao
\thanks{W.B. Du, W. Ying, Y.B. Zhu and X.B. Cao are with Beijing Key Laboratory for Network-based Cooperative Air Traffic Management, the School of Electronic and Information Engineering, Beihang University, Beijing 100191, P.R.China (e-mail: wenbodu@buaa.edu.cn; yingwen\_92@163.com; yanbo\_zhu@163.com; xbcao@buaa.edu.cn).}
\thanks{G. Yan is with Center for Complex Network Research and Department of Physics, Northeastern University, Boston, Massachusetts 02115, USA (e-mail: eegyan@gmail.com).}
\thanks{Manuscript received April 19, 2016; revised.}}

%
%

\markboth{Ieee transactions on circuits and systems---II:express briefs}%
{Ying \MakeLowercase{\textit{et al.}}: Heterogeneous Strategy Particle Swarm Optimization}
%



\maketitle

\begin{abstract}
PSO is a widely recognized optimization algorithm inspired by social swarm. In this brief we present a heterogeneous strategy particle swarm optimization (HSPSO), in which a proportion of particles adopt a fully informed strategy to enhance the converging speed while the rest are singly informed to maintain the diversity. Our extensive numerical experiments show that HSPSO algorithm is able to obtain satisfactory solutions, outperforming both PSO and the fully informed PSO. The evolution process is examined from both structural and microscopic points of view. We find that the cooperation between two types of particles can facilitate a good balance between exploration and exploitation, yielding better performance. We demonstrate the applicability of HSPSO on the filter design problem.
\end{abstract}


%
\IEEEpeerreviewmaketitle

\section{Introduction}
\IEEEPARstart{P}{ARTICLE} swarm optimization (PSO) is a typical swarm intelligence optimization algorithm inspired by animal social behaviors, such as bird flocking and fish schooling \cite{1}. A group of particles in PSO fly in the search space, aiming to find the optimum cooperatively. Each particle exchanges information with others and learns useful information to improve its performance. Due to its ease to implement and outstanding performance, PSO has been widely used to solve real-world schedule or engineering problems such as antennas\cite{app1}, system control\cite{app2}, electronics and electromagnetics \cite{app3}.

In the original PSO \cite{1}, each particle learns from the best historical experience of the whole population. The concepts of structure and neighbors in PSO were first introduced in \cite{5}, where each particle learns from the best historical experience of its neighbors. However, in both versions the particles learn from the best individual, hence some useful information of other individuals is neglected \cite{5}. To take the advantage of full information, the fully informed particle swarm optimization (FIPSO) was proposed \cite{6} where all neighbors are information sources. Though FIPSO can rapidly converge, it may miss some promising regions in the search space of the optimization problem.

Most previous works treated all individuals as the same, neglecting the individual heterogeneity. Actually, the individual heterogeneity plays an important role in swarm intelligence and has been verified to be able to significantly improve the performance of PSO \cite{7,8}.
Here, we propose a heterogeneous strategy particle swarm optimization (HSPSO), in which a proportion of particles are single informed, while others are fully informed. Our experimental results show that HSPSO obtains satisfactory solutions and outperforms FIPSO and canonical singly informed PSO (SIPSO), because in HSPSO fully-informed particles can adequately utilize the global information and guide the swarm while singly-informed particles can maintain the diversity.

The rest of the paper is organized as follows. Section II introduces HSPSO in detail and shows its relation to SIPSO and FIPSO. Section III compares the results of three PSOs. Section IV employs HSPSO to solve the problem of 2-Dimensional recursive filter design. Section V makes a conclusion.

\section{Algorithm Description}

In HSPSO, $N$ particles fly in a $D$-dimensional space to search the optimum. The $i$th particle updates its velocity and position of $d$th dimension by

\begin{align}
&x_{i}^d:=x_i^d+v_i^d\\
&v_{i}^d:=\chi[v_{i}^d+\frac{\varphi}{2}r_1(p_i^d-x_i^d)+\frac{\varphi}{2}r_2(p_{i_{nb}}^d-x_i^d]\\
&v_{i}^d:=\chi[v_{i}^d+\frac{\varphi}{k_i}\sum_{m=1}^{k_i}r_{m}(p_{i_m}^d-x_{i}^d)]
\end{align}

where (2) is for singly-informed (SI) particles while (3) is for fully-informed (FI) particles, $\varphi$=4.1 and $\chi$=0.729 according to common practices \cite{6,7,8,LI}, $p_i=[p_i^1,p_i^2,...,p_i^D]$ denotes the historical best position of particle $i$, $p_{i_{nb}}=[p_{i_{nb}}^1,p_{i_{nb}}^2,...,p_{i_{nb}}^D]$ denotes the historical best position in all neighbors of particle $i$, $k_i$ is the number of the $i$th particle's neighbors, $i_m$ is the $m$th neighbor of the particle $i$, $p_{i_m}=[p_{i_m}^1,p_{i_m}^2,...,p_{i_m}^D]$ is the historical best position of $i_m$, $r_1$, $r_2$ in (2) and all of $r_m$ in (3) are independent random numbers in range $[0,1]$.

\begin{figure}[!t]
\centering
    \includegraphics[width=0.45\textwidth]{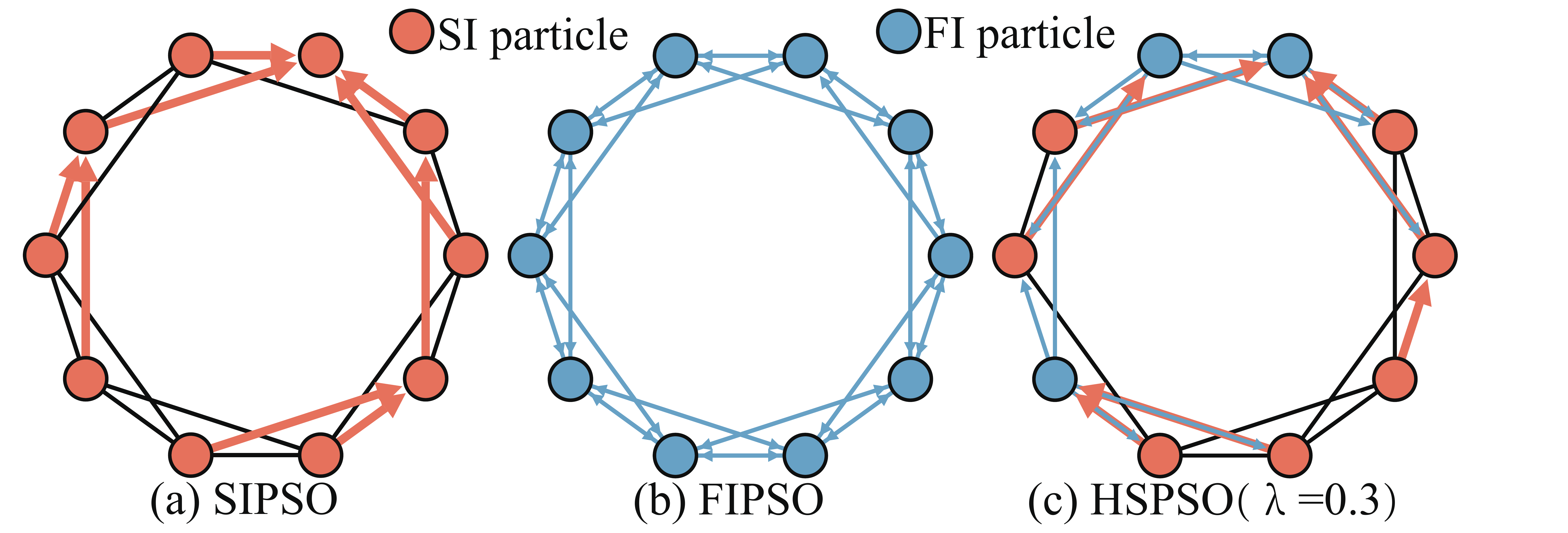}
    \caption{Schematic diagram of three PSOs. Each directed edge shares the same color with its source, denoted the source particle learn from the target particle. Black edges indicate the links without information interaction.}\label{F1}
\end{figure}

Note that there are two strategies of updating the velocity, i.e. FI and SI. Each particle employs the alternative velocity formula according to its property. Here we use a parameter $\lambda \in [0, 1]$ to divide the swarm into two groups. A group of particles, with size $\lfloor \lambda N \rfloor$, are randomly selected as fully-informed (FI) particles, and the rest are singly-informed (SI) ones. Fig. \ref{F1} illustrates this feature, where a widely used ring structure with average degree $\overline{k}$=4 is employed for instance. One can see that each particle is influenced by the best one of $4$ neighbors in SIPSO Fig. \ref{F1}(a), and by all of 4 in FIPSO Fig. \ref{F1}(b), while both types exist in HSPSO Fig. \ref{F1}(c) with a certain proportion ($\lambda$=0.3 in this example) of individuals.

The $\lambda$ is a key parameter in HSPSO algorithm to balance the effect of FI and SI particles, because superfluous FI particles could provide too much redundant information while an excess of SI particles may result in information loss. Specifically, when $\lambda$=1, all particles are FI particles, then HSPSO degrades to FIPSO. When $\lambda$=0, HSPSO becomes SIPSO.

\section{Expermental Results}
\subsection{Test Functions and Conditions}

To evaluate the performance of HSPSO, we employ six widely-used benchmark functions \cite{8,9}. The formulas and the details of these functions are listed in Table \ref{T1}. Among these functions, $f_1$(sphere), $f_2$(Rosenbrock) and $f_3$(Quartic Noise) are unimodal function, yet $f_2$ is sometimes treated as multimodal function when $D$ is large, and $f_3$ includes a stochastic term. The other $3$ functions are multimodal, where $f_4$(Ackley) is the simplest one, while the landscape of $f_5$(Rastrigin) is more complex with many deep local optima, and $f_6$(Griewank) are asymmetrical. The dimension of all these benchmark functions are set as $D=30$. With such diverse characteristics, these functions could help test the performance of HSPSO in a comprehensive way.

The rest experiments adopt the following parameter setting: the population size $N=50$, each run stops at $5000$ iterations and each data is averaged by $100$ times.

\begin{table}[!t]
\centering
\caption{Benchmark Functions}
\begin{tabular}{|p{6cm}<{\centering}|c|}
\hline
Formula & Range\\
\hline
$f_1(x)=\sum_{i=1}^D{{x_i}^2}$ & $[-100,100]^D$\\
\hline
$f_2(x)=\sum_{i=1}^{D-1}{100(x_{i+1}-{x_i}^2)^2+(x_i-1)^2}$ & $[-30,30]^D$\\
\hline
$f_3(x)=\sum_{i=1}^{D}{ix_i^4+random[0,1)}$ & $[-1.28,1.28]^D$\\
\hline
$f_4(x)=-20exp(-0.2\sqrt{\frac{1}{D}\sum_{i=1}^{D}{x_i^2}})-exp(-0.2\sqrt{\frac{1}{D}\sum_{i=1}^{D}{cos2\pi x_i}})+20+e$ & $[-32,32]^D$\\
\hline
$f_5(x)=\sum_{i=1}^D{{x_i}^2-10\cos{2\pi{x_i}}+10}$ & $[-5.12,5.12]^D$\\
\hline
$f_6(x)=\frac{1}{4000}\sum_{i=1}^D{{x_i}^2}-\prod_{i=1}^D{\cos{\frac{x_i}{\sqrt{i}}}}+1$ & $[-600,600]^D$\\
\hline
\end{tabular}
\label{T1}
\end{table}

\subsubsection{Algorithm Performances}

\begin{figure}[!t]
\centering
    \includegraphics[width=0.45\textwidth]{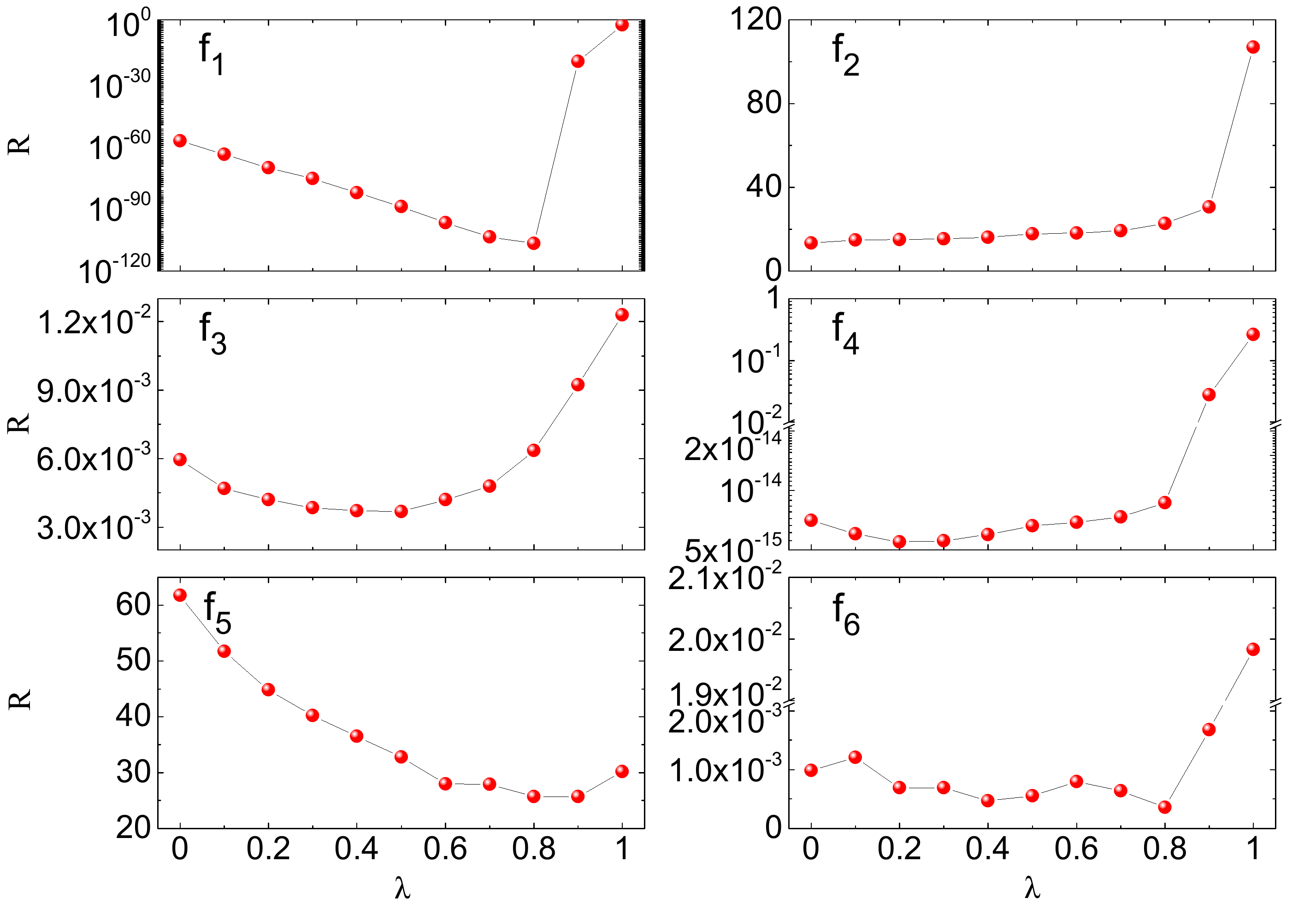}
    \caption{Solution quality $R$ with variation of $\lambda$.}
    \label{F2}
\end{figure}

We compare the performance of HSPSO to that of SIPSO and FIPSO, i.e. HSPSO with $\lambda=0$ and with $\lambda=1$ under the criteria of solution quality $R$ (the final optimized fitness value), which is the most important criteria. Firstly, we investigate $R$ of the algorithm under the ring structure with $\overline{k}$=4, where $\lambda$ varies from $0$ to $1$. As is shown in Fig. \ref{F2}, HSPSO with an appropriate $\lambda$ outperforms both canonical PSO and FIPSO on almost all of test functions. Moreover, $\lambda$ is various with different functions. It reveals that the cooperation of FI particles and SI particles helps to improve the optimization process under an appropriate proportion of FI particles.

\begin{figure}[!t]
\centering
    \includegraphics[width=0.45\textwidth]{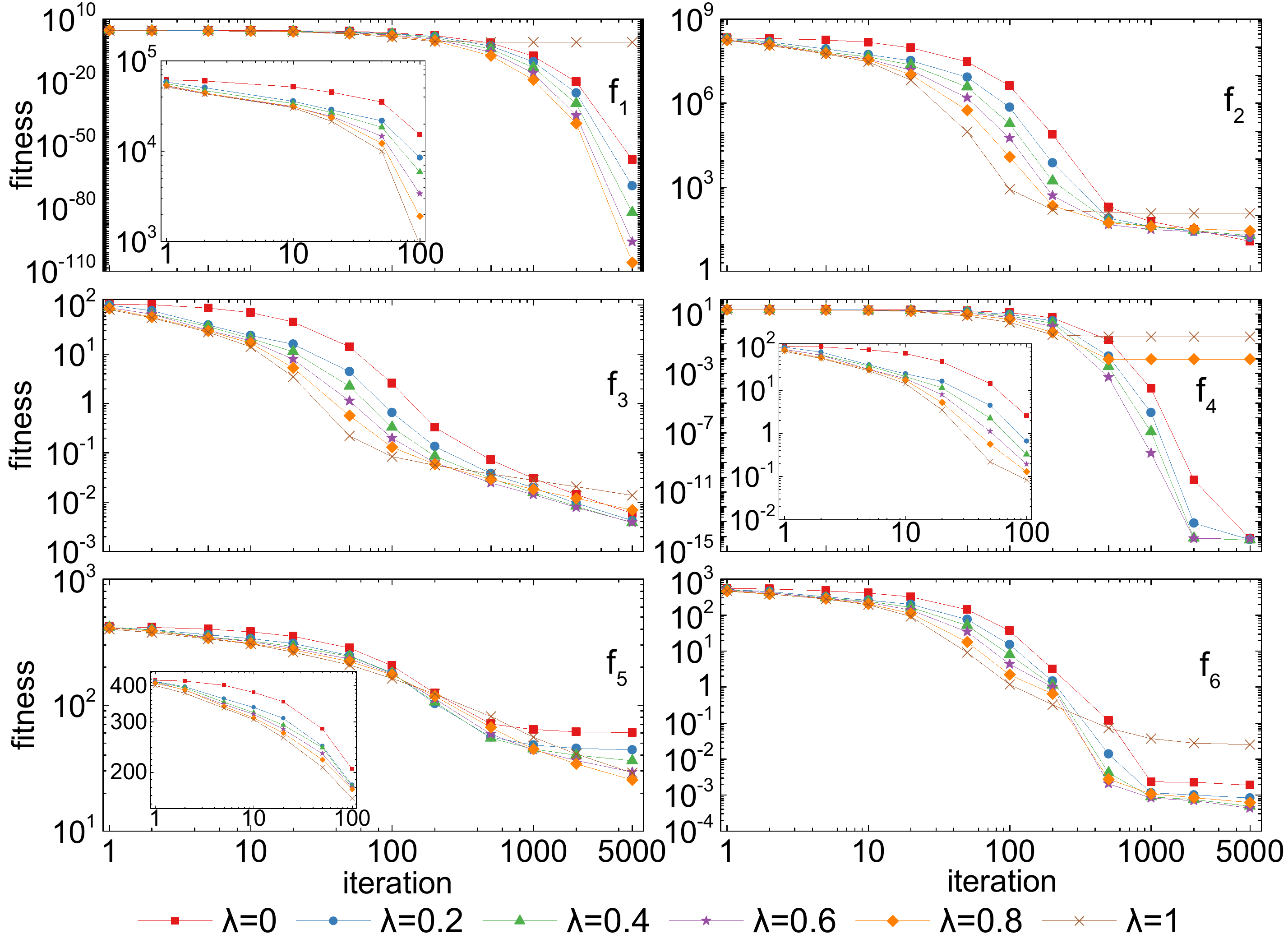}
    \caption{Fitness variation during optimization process.}
    \label{F3}
\end{figure}

To investigate the optimization process in more details, we examine the variation of fitness value during the evolution. As shown in Fig. \ref{F3}, HSPSO with a larger $\lambda$, especially FIPSO, converges faster than HSPSO with small $\lambda$ at the beginning of the evolution. However, the premature convergence will make the swarm stagnate, not finding more promising solutions. Hence the $R$ value of FIPSO is usually unsatisfactory. Canonical PSO is rarely troubled by premature, yet it converges quite slowly. HSPSO is outstanding because FI particles could ensure an appropriate convergence speed, while SI particles maintain the diversity of the swarm. Therefore, HSPSO with an appropriate $\lambda$ could converge faster than canonical PSO and avoid premature meanwhile.

\subsection{The Impact of Topology}

\begin{figure}[!t]
\centering
    \includegraphics[width=0.45\textwidth]{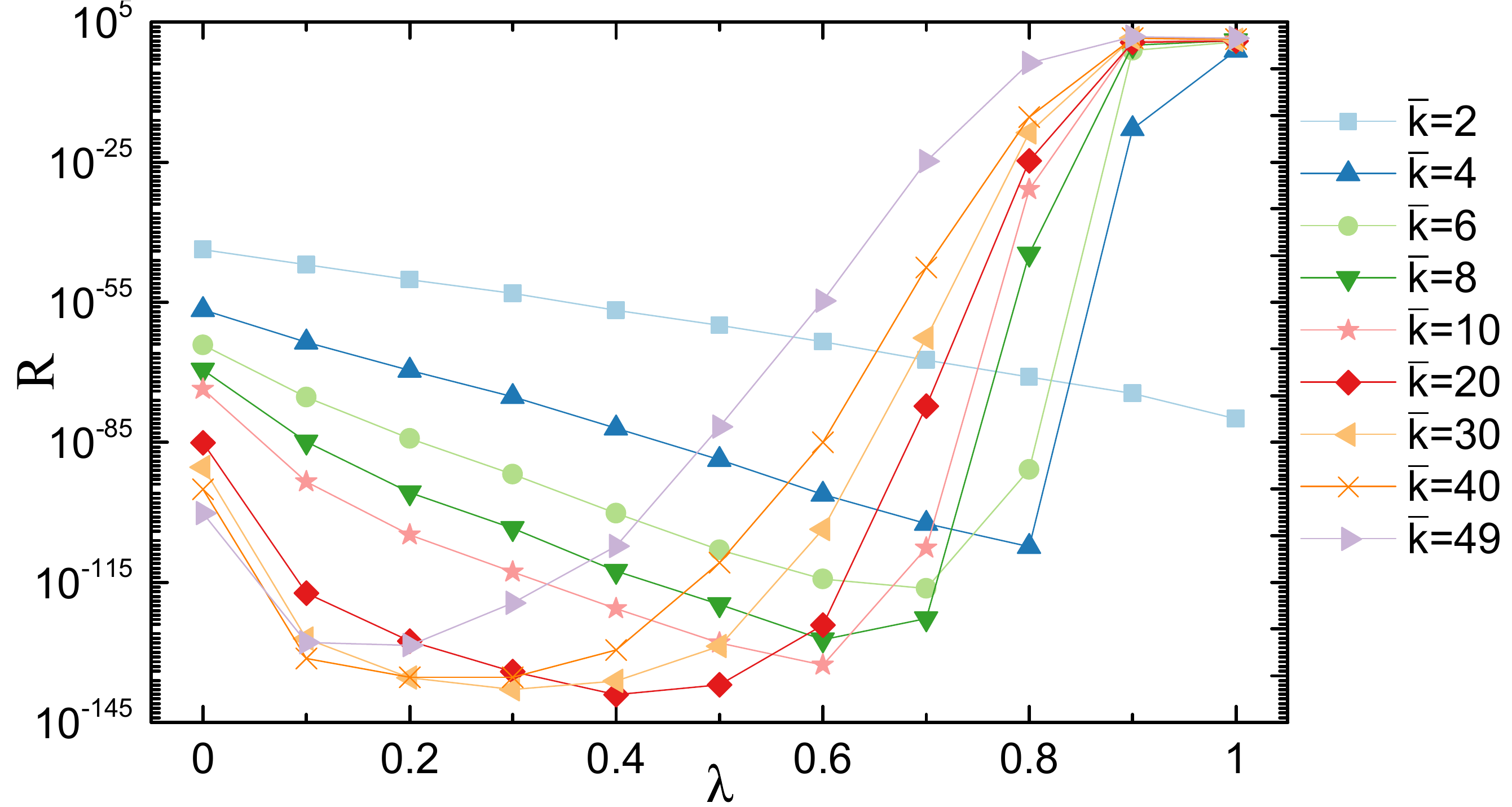}
    \caption{Solution quality $R$ of HSPSO with different network sparsity and variation of $\lambda$ to solve $f_1$.}
    \label{F4}
\end{figure}

A key advance in understanding complex networks over the last decade has been how powerfully network topology affects many network properties and dynamical processes \cite{netw1}-\cite{netw4}. Though the idea of HSPSO is mainly about learning strategy, the topology is also an important factor. As a network-based information system, PSO's performance is greatly influenced by the network sparsity. A dense network makes information spread fast. Yet a network with a small average degree impedes the information spreading, in which particles could preferably maintain the diversity. Thus, we further investigate the impact of topology sparsity. As shown in Fig. \ref{F4}, the optimal $\lambda$, inducing best $R$, decreases with the increase of $\overline{k}$. In a dense network, FI particles speed up the process of spreading information due to the abundant neighbors, which may lead to premature convergence. Plenty of FI particles which absorb information without discrimination will weaken valuable information, even mislead each other, while the mechanism of SI particle could discriminate information effectively. Therefore, to avoid confusion, the better choice is to employ much fewer FI particles than SI ones in a dense network.

\begin{figure}[!t]
\centering
    \includegraphics[width=0.45\textwidth]{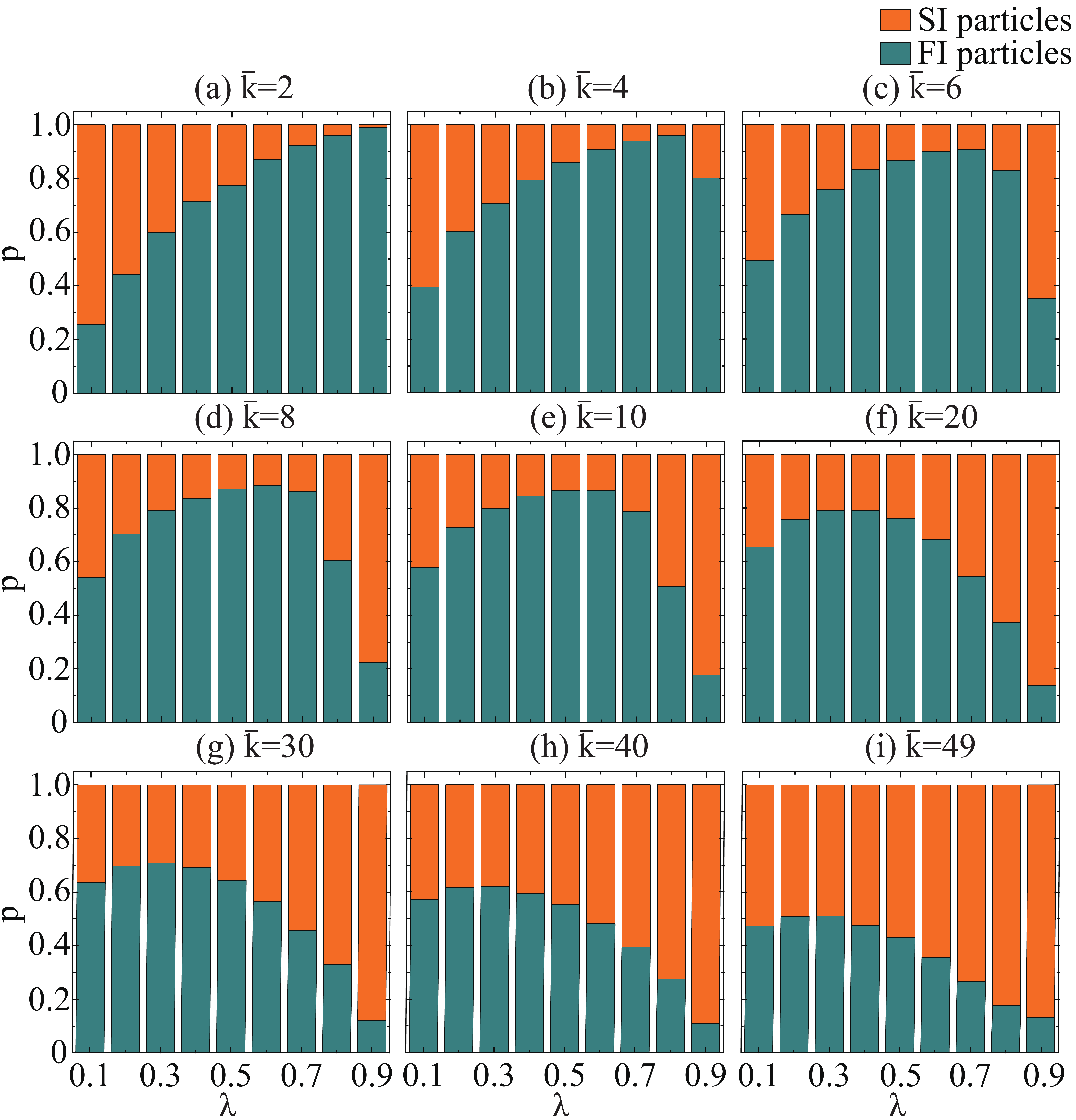}
    \caption{Percentage $p$ that FI particles discover better solutions during the whole evolution iterations. $p=num_{FI}/num_{total}$, where $num_{FI}$ is the number that FI particles find better solutions and $num_{total}$ is the total number that all particles find better solutions.}
    \label{F5}
\end{figure}

To further uncover the underlying mechanism of the optimization process, we examine the exploring ability of FI particles. In Fig. \ref{F5}, $p$ is denoted as the percentage that FI particles discover better solutions. Interestingly, the optimal $\lambda$ in Fig. \ref{F4} is well consistent with the maximal $p$ in Fig. \ref{F5}, indicating that the performances of FI particles are evidently relevant to the solution quality of HSPSO. In other words, FI particles are more likely to act as guiders in the swarm due to the FI learning strategy. Furthermore, as $\overline{k}$ increases, the appropriate $\lambda$ for the maximum of $p$ decreases, implying that fewer guiders are needed to lead the swarm in more densely-connected networks. When $\lambda$ is small, the minority FI particles are powerless while SI particles which are adept at maintaining the diversity can not use information effectively. If $\lambda$ is too large, on the contrary, the redundant information will mislead FI particles, thus SI particles will play an effective role to pull the swarm out of a local optimum.

\begin{figure}[!t]
\centering
    \includegraphics[width=0.45\textwidth]{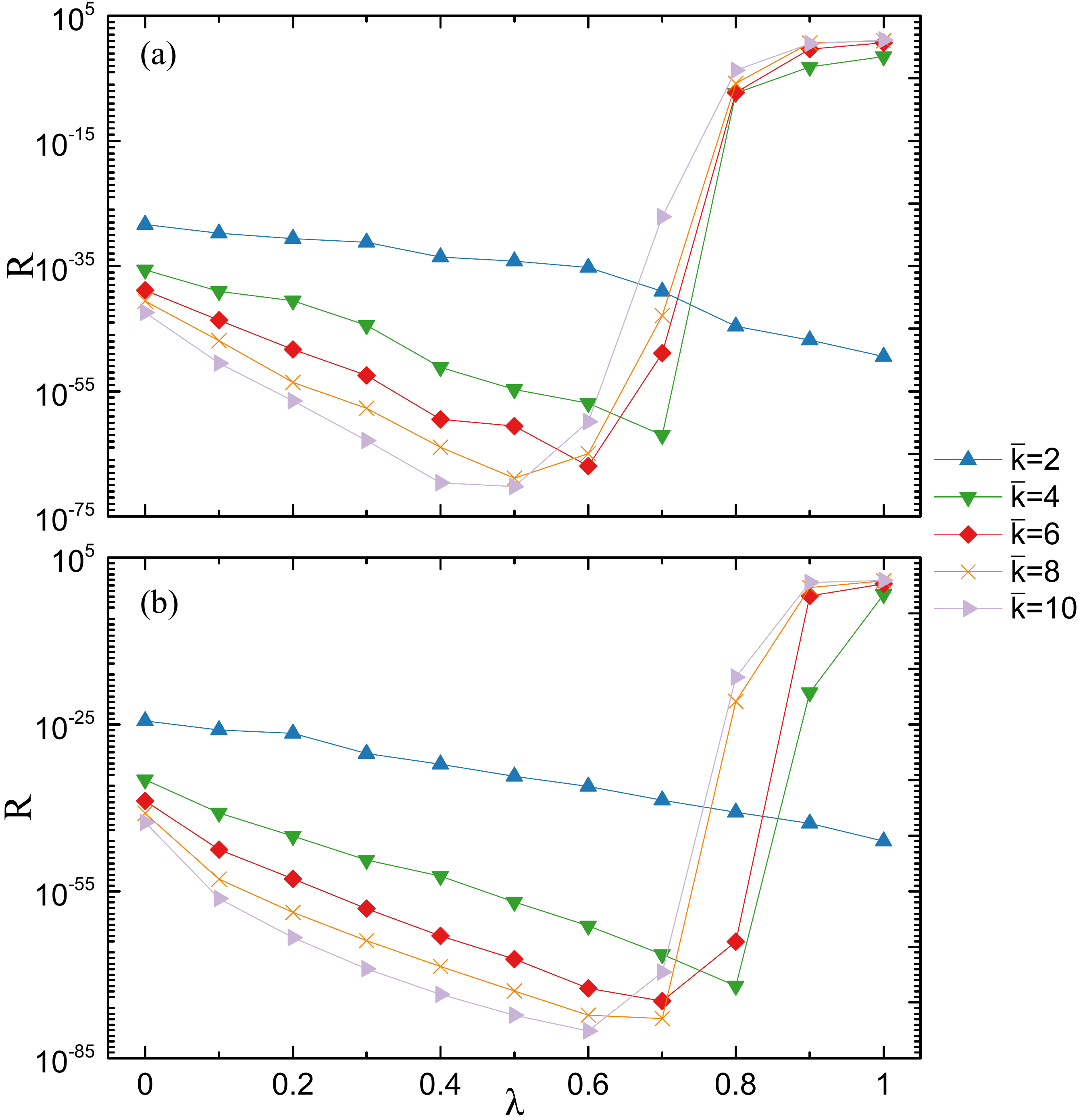}
    \caption{Solution quality $R$ vs $\lambda$ when HSPSO employs a scale-free network (a) or a small-world network (b).}
    \label{F6}
\end{figure}

We also investigate other networks, such as scale free network \cite{sf} (in Fig. \ref{F6}(a)) and small world network \cite{sw} (in Fig. \ref{F6}(b)). In consideration of the appropriateness of network sparsity, $\overline{k}$ of these networks are set no more than 10. As expected, in Fig. \ref{F6}, HSPSO with these topologies show similar results to Fig. \ref{F4}, demonstrating the robustness of our algorithm. Futhermore, some relatively novel network structures such as in \cite{other-net1} and \cite{other-net2} will be investigated in our future work.

\section{HSPSO for the Design of Two-Dimensional IIR Digital Filters}
\subsection{Problem Description}
To demonstrate the applicability of HSPSO we use it to solve a design problem of IIR digital filters, which attracted considerable attentions during past decades \cite{IIR1}-\cite{IIR4}.

The transfer function of $2$-D recursive digital filters can be described by
\begin{equation}
H(z_1,z_2)=H_0\frac{\sum_{i=0}^{N}{\sum_{j=0}^{N}{a_{ij}z_1^iz_2^j}}}{\prod_{l=1}^{N}{1+b_lz_1+c_lz_2+d_lz_1z_2}}, a_{00}=1
\end{equation}
where $N$ is the dimension of the filter, $z_1=e^{-j\omega_1}$ and $z_2=e^{-j\omega_2}$, and $\omega_1$, $\omega_2$ are the frequencies in range $[-\pi,\pi]$. The task of filter designing is to adjust the coefficients of $M(\omega_1,\omega_2)=H(z_1,z_2)$ to approximate the desired amplitude response of the $2$-D filter $M_d(\omega_1,\omega_2)$. In this brief, the desired amplitude response $M_d(\omega_1,\omega_2)$ follows \cite{GA} as
\begin{equation}
M_d(\omega_1,\omega_2)=\left\{ \begin{aligned}
& 1, && \sqrt{\omega_1^2+\omega_2^2}\leqslant0.08\pi &\\
& 0.5, && 0.08\pi<\sqrt{\omega_1^2+\omega_2^2}\leqslant0.12\pi &\\
& 0, && \sqrt{\omega_1^2+\omega_2^2}>0.12\pi & \end{aligned} \right.
\end{equation}
Hence, the design of $2$-D filter can be formalized as an optimization problem of minimizing the cost function
\begin{equation}
\begin{aligned}
J_p & = J(a_{ij},b_l,c_l,d_l,H_0)\\
& =\sum_{l_1=0}^{N_1}{\sum_{l_2=0}^{N_2}{\left[\left|M(\frac{\pi l_1}{N_1},\frac{\pi l_2}{N_2})-M_d(\frac{\pi l_1}{N_1},\frac{\pi l_2}{N_2})\right|\right]^p}}
\end{aligned}
\end{equation}
s.t.
\begin{equation}
|b_l+c_l|-1<d_l,\quad d_l<1-|b_l-c_l|,\quad l=1,2,...N.
\end{equation}
where $p=2$, and $N_1=N_2=50$. The cost function describes the difference of $M(\omega_1,\omega_2)$ and $M_d(\omega_1,\omega_2)$ in $N_1\times N_2$ points.

\subsection{Experimental Results}

\begin{figure*}[!t]
\centering
    \includegraphics[width=0.9\textwidth]{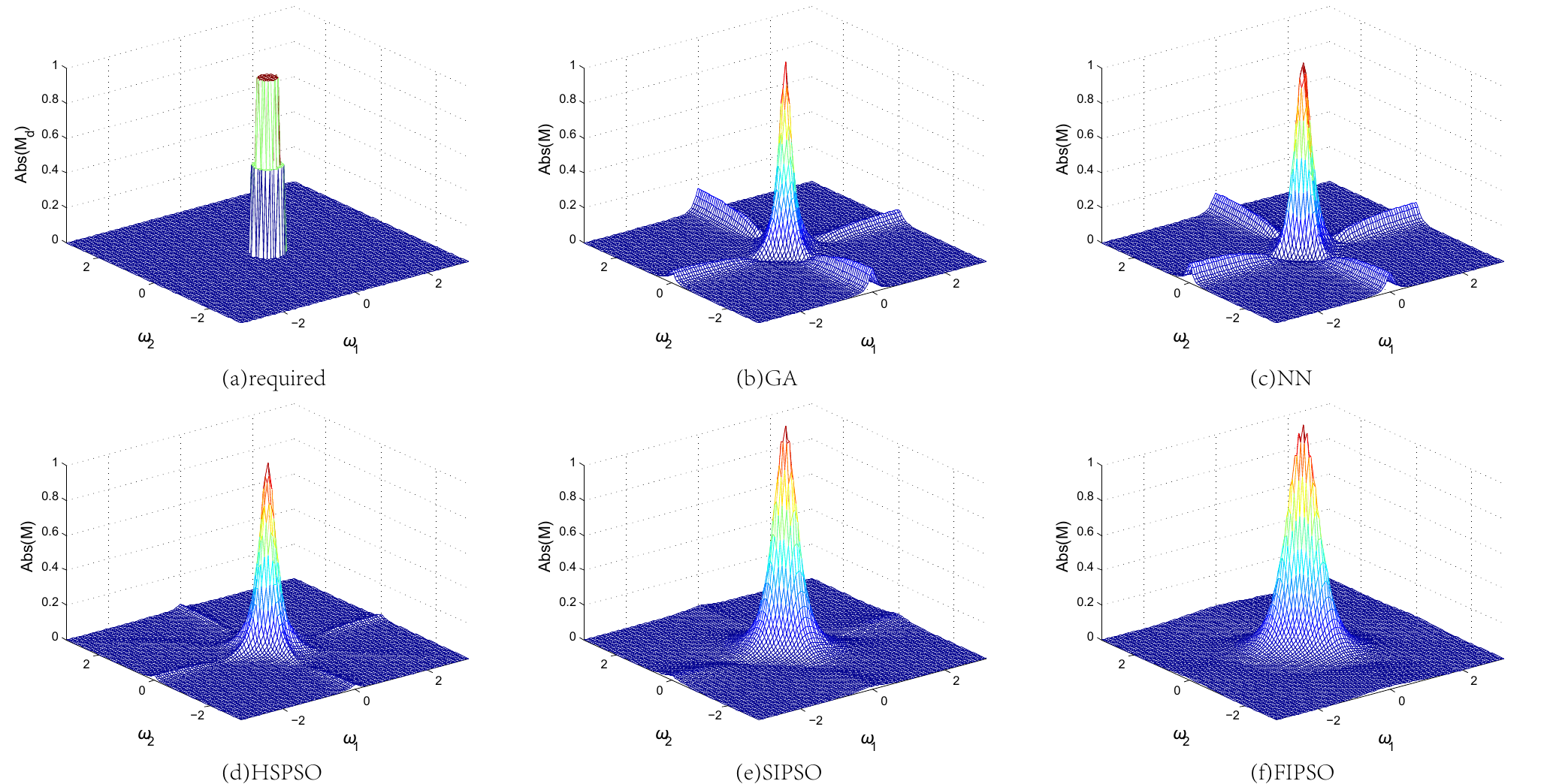}
    \caption{Amplitude frequency response of the (a) required 2-D filter $|M_d(\omega_1,\omega_2)|$, (b)-(f) 2-D filter $|M(\omega_1,\omega_2)|$ designed by different algorithms.}
    \label{F7}
\end{figure*}

As $J_p$ is the function of $a_{ij}$, $b_l$, $c_l$ and $H_0$, we construct a vector $x$=[$a_{01}$, $a_{02}$, $a_{10}$, $a_{11}$, $a_{12}$, $a_{20}$, $a_{21}$, $a_{22}$, $b_1$, $b_2$, $c_1$, $c_2$, $d_1$, $d_2$, $H_0$] for HSPSO. The parameter of HSPSO is set as follow: population size $N$ is set as $50$, all variables in vector $x$ are in the range of $[-3,3]$ \cite{GA,NN}, the evolution lasts for $2000$ iterations. Table \ref{T2} lists the parameters optimized by HSPSO and other competitors, including Genetic Algorithm (GA)\cite{GA}, Neural Network (NN)\cite{NN}, SIPSO (HSPSO with $\lambda=0$, $\overline{k}=2$) and FIPSO (HSPSO with $\lambda=1$, $\overline{k}=2$).

Fig. \ref{F7} shows the frequency response of the required filter and the designed filters with the parameters in Table \ref{T2}. One can see that HSPSO performs better than GA and NN methods. Note that the high frequency region of filters designed by SIPSO and FIPSO are flat, yet the low frequency region are not satisfactory, due to its elliptical transverse section rather than a circle. Therefore, HSPSO outperforms both SIPSO and FIPSO due to the cooperation of singly- informed particles and fully-informed particles.

\begin{table}[!t]
\centering
\caption{The results of optimized filter coefficients}
\begin{tabular}{|c|c|c|c|c|c|}
\hline
parameters & NN & GA & SIPSO & FIPSO & HSPSO\\
\hline
$a_{01}$ & 1.8922 & 1.8162 & 0.3801 &-0.0380 &-2.104\\
\hline
$a_{02}$ &-1.2154 &-1.1060 & 0.2545 & 0.5724 &-1.5145\\
\hline
$a_{10}$ & 0.0387 & 0.0712 &-0.1083 & 0.6357 &-2.2828\\
\hline
$a_{11}$ &-2.5298 &-2.5132 & 0.4721 &-0.4270 & 2.7886\\
\hline
$a_{12}$ & 0.3879 & 0.4279 &-0.8995 & 0.3376 & 1.5839\\
\hline
$a_{20}$ & 0.6115 & 0.5926 & 0.5398 & 0.7397 &-1.2061\\
\hline
$a_{21}$ &-1.4619 &-1.3690 &-1.2448 &-0.0664 & 1.1080\\
\hline
$a_{22}$ & 2.5206 & 2.4326 & 2.3634 & 1.2504 &-2.7257\\
\hline
$b_1$    &-0.8707 &-0.8662 &-0.7536 &-0.4355 &-0.9260\\
\hline
$b_2$    &-0.8729 &-0.8907 &-0.3749 &-0.4537 &-0.4123\\
\hline
$c_1$    &-0.8705 &-0.8531 &-0.7789 &-0.5386 &-0.9376\\
\hline
$c_2$    &-0.8732 &-0.8388 &-0.4028 &-0.3609 &-0.2998\\
\hline
$d_1$    & 0.7756 & 0.7346 & 0.5816 & 0.0791 & 0.8846\\
\hline
$d_2$    & 0.7799 & 0.8025 &-0.1003 &-0.0694 &-0.1859\\
\hline
$H_0$    & 0.0010 & 0.0009 & 0.0028 & 0.0039 & 0.0007\\
\hline
\end{tabular}
\label{T2}
\end{table}
\section{Conclusion}
In this brief we propose HSPSO, a swarm optimization algorithm composed of two types of particles with different learning strategies. We test the performance of HSPSO on six widely-used benchmark functions. Our results show that HSPSO is superior to canonical PSO and FIPSO. Our investigation on the impact of network topology and the underlying mechanism of HSPSO reveals that the heterogeneity of the swarm results in the division and cooperation between different particles, leading to a more effective optimization process. The successful application of HSPSO to $2$-D filter design problem demonstrates its applicability in solving real-world optimization problems.


%



\section*{Acknowledgment}
This paper is supported by the National Natural Science Foundation of China (Grant Nos. 61425014, 61521091), National Key Research and Development Program of China (Grant No. 2016YFB1200100), and National Key Technology R\&D Program of China (Grant No. 2015BAG15B01).

\ifCLASSOPTIONcaptionsoff
  \newpage
\fi

\end{document}